\documentclass[a4paper,twoside]{article}
\pdfoutput=1

\usepackage{graphicx} 
\graphicspath{{./figures/}}
\usepackage{caption,subcaption}
\usepackage{scalerel}
\usepackage[table]{xcolor}

\usepackage{times}
\usepackage{lipsum}  
\usepackage{mathtools}
\usepackage{amsmath}
\usepackage{amssymb}
\usepackage{tabularx}
\usepackage{tabularx}
\usepackage{url}
\usepackage[utf8]{inputenc}
\usepackage[T1]{fontenc}
\usepackage{textcomp}
\usepackage{gensymb}
\usepackage{multirow}
\usepackage{booktabs}
\usepackage{etoolbox}
\usepackage{adjustbox}
\usepackage[noadjust]{cite}
\usepackage{soul}
\usepackage{float}
\usepackage{amsmath}
\usepackage{algorithm2e}
\usepackage[noend]{algpseudocode}
\newfloat{algorithm}{t}{lop}
\usepackage{cite}
\usepackage{multicol}
\usepackage{leftidx}
\usepackage[english]{babel}
\usepackage{blindtext}
\usepackage{epsfig}

\usepackage{SCITEPRESS}

\newcommand{\change}[1]{\textcolor{black}{#1}}
\usepackage[bookmarks=true]{hyperref}
\usepackage{ifthen} 
\newboolean{preprint} 
\setboolean{preprint}{true} 
\ifthenelse{\boolean{preprint}}%
    {
        \newcommand{\PREPRINTYEAR}{2025}
        \newcommand{\PUBLISHEDINSHORT}{ICINCO 2025}
        \newcommand{\PUBLISHEDIN}{International Conference on Informatics in Control, Automation and Robotics  (\PUBLISHEDINSHORT)}
        \newcommand{\DOI}{XX.XXXX/XXX.XXXX.XXXXXX} 
        
        \usepackage[placement=top,vshift=-7]{background}
        \SetBgScale{1.0}
        \SetBgContents{\PUBLISHEDINSHORT. PREPRINT VERSION - DO NOT DISTRIBUTE. \href{https://doi.org/\DOI}{DOI \DOI}}
        \SetBgColor{black}
        \SetBgAngle{0}
        \SetBgOpacity{1.0}
    }

\begin{document}
\ifthenelse{\boolean{preprint}}%
    {
        \thispagestyle{empty}
        \onecolumn
        {
          \topskip0pt
          \vspace*{\fill}
          \centering
          \LARGE{%
            \copyright{} \PREPRINTYEAR~\PUBLISHEDIN\\\vspace{1cm}
            Personal use of this material is permitted.
            Permission from \PUBLISHEDIN \footnote{\url{https://icinco.scitevents.org/}}~must be obtained for all other uses, in any current or future media, including reprinting or republishing this material for advertising or promotional purposes, creating new collective works, for resale or redistribution to servers or lists, or reuse of any copyrighted component of this work in other works.}
            \vspace*{\fill}
        }
        \NoBgThispage
        \twocolumn          	
        \BgThispage
    }

\title{Simultaneous learning of state-to-state minimum-time planning and control}

\author{
\authorname{
Swati Dantu\orcidAuthor{0009-0003-8895-3462}, 
Robert Pěnička\orcidAuthor{0000-0001-8549-4932} and 
Martin Saska\orcidAuthor{0000-0001-7106-3816}
}
\affiliation{Department of Cybernetics, Faculty of Electrical Engineering, Czech Technical University in Prague} \email{swati.dantu@fel.cvut.cz}
}

\abstract{
This paper tackles the challenge of learning a generalizable minimum-time flight policy for UAVs, capable of navigating between arbitrary start and goal states while balancing agile flight and stable hovering.
Traditional approaches, particularly in autonomous drone racing, achieve impressive speeds and agility but are constrained to predefined track layouts, limiting real-world applicability. 
To address this, we propose a reinforcement learning-based framework that simultaneously learns state-to-state minimum-time planning and control and generalizes to arbitrary state-to-state flights. 
Our approach leverages Point Mass Model (PMM) trajectories as proxy rewards to approximate the true optimal flight objective and employs curriculum learning to scale the training process efficiently and to achieve generalization. 
We validate our method through simulation experiments, comparing it against Nonlinear Model Predictive Control (NMPC) tracking PMM-generated trajectories and conducting ablation studies to assess the impact of curriculum learning. 
Finally, real-world experiments confirm the robustness of our learned policy in outdoor environments, demonstrating its ability to generalize and operate on a small ARM-based single-board computer.}

\onecolumn \maketitle \normalsize \setcounter{footnote}{0} \vfill

\section{\uppercase{Introduction}}
Unmanned Aerial Vehicles (UAVs) are becoming increasingly used in applications that benefit from fast and agile flight between given goals, such as inspection~\cite{Hwei-Ming2021WindFarmInspection}, surveillance~\cite{Hsiang-Chun2023surveillance}, and humanitarian search and rescue~\cite{Alotaibi2019SAR}. 
In these scenarios, the ability to fly between desired positions as quickly as possible is crucial for maximizing efficiency. 
The problem of flying a drone to a target location in minimum time typically involves two key aspects: (1) planning a minimum-time trajectory and (2) controlling the drone such that it accurately follows the planned trajectory. 
Successfully solving this problem allows UAVs to navigate at high speeds and thus efficiently finish e.g. an inspection mission.

Solving for planning and control together while achieving minimum-time flight is, however, challenging due to several fundamental reasons. 
UAVs have highly nonlinear dynamics and must be controlled at the edge of their actuation limits to achieve the fastest possible motion. 
Unlike stable hovering or slow navigation, agile flight demands precise control under extreme conditions, where any model mismatches or external disturbances can lead to catastrophic failures. 
A significant challenge remains in designing an algorithm, in our case a learning-based planner-controller leveraging Deep Reinforcement Learning (DRL), that can generalize to arbitrary starting and ending states, and thus can solve simultaneously the state-to-state planning and control for general purpose flight.
This is in contrast, e.g., with existing learning-based approaches for UAV racing~\cite{kaufmann2023champion} that have been designed to work only with predefined gate sequences.

\begin{figure}[t!]
	\centering
	\includegraphics[width=1.0\columnwidth]{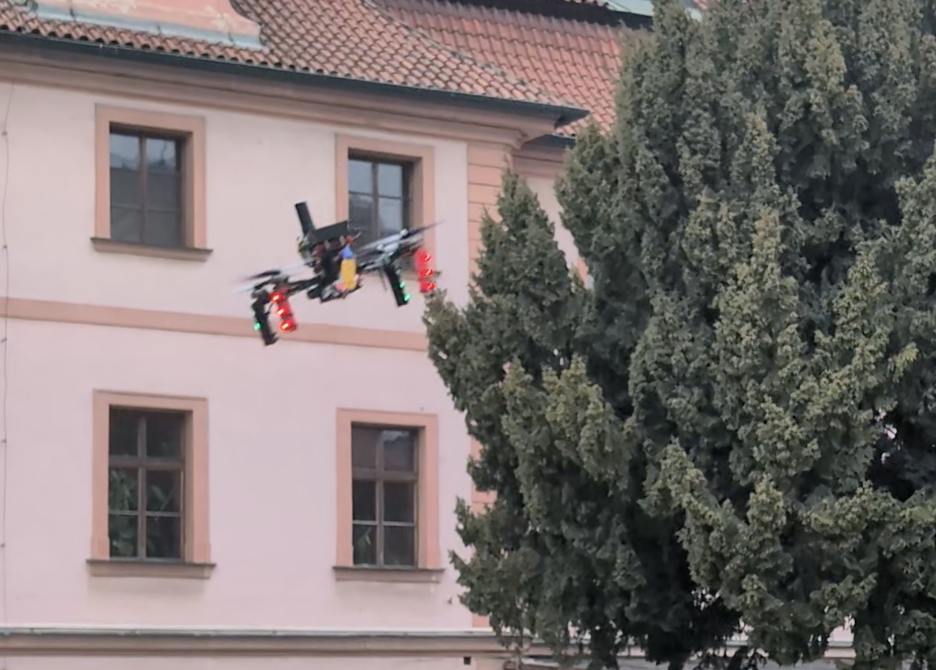}
	\caption{Illustration from real-world flight: quadrotor in flight using proposed learned policy. }
	\label{fig:intro_picture}
    \vspace{-1em}
\end{figure}

Prior work on learning-based UAV flight has primarily focused on specific racing scenarios~\cite{kaufmann2023champion}, achieving impressive performance but failing to generalize beyond small deviations in predefined tracks~\cite{song2021autonomous}. 
While recent advances demonstrate the ability to learn flight policies in a matter of seconds~\cite{Eschmann_2024_learning_in_seconds} or adapt to different UAV configurations~\cite{zhang2024learningbasedquadcoptercontrollerextreme}, these approaches have been validated only under near-hover conditions or moderate speeds. 
The key limitation remains the lack of a robust, learning-based minimum-time planner-controller that can generalize to arbitrary start-goal positions while maintaining agility in general flight conditions.

To bridge this gap, we propose a  approach for learning-based minimum-time flight by effectively solving both planning and control with a single policy that generalizes to arbitrary start-goal configurations. 
We leverage Deep Reinforcement Learning approach that incorporates two key ideas: (1) leveraging Point Mass Model (PMM) trajectories as proxy rewards, ensuring that the learning objective closely approximates the true performance metric of minimum-time flight~\cite{matej_2024_pmm}, and (2) employing a curriculum learning strategy to progressively scale the learning process, enabling the policy to handle increasingly large position variations in agile flight and at the same time be able to hover smoothly.
To the best of our knowledge, this is the first work that integrates these concepts into a reinforcement learning framework for UAV minimum-time flight, achieving a balance between hovering stability and high-speed manoeuvrability.

We evaluate the proposed approach in both simulation and real-world experiments. 
In simulation, we compare our learned policy against Nonlinear Model Predictive Control (NMPC) tracking PMM-generated trajectories over various maneuvers, including straight-line, zigzag, and semicircular paths, demonstrating comparable flight times while maintaining more stable execution. 
We show that the curriculum learning significantly improves convergence and performance compared to conventional reinforcement learning. 
Finally, real-world tests validate our policy's robustness in outdoor environments, showcasing its ability to control a small agile drone with limited compute power as shown in Figure~\ref{fig:intro_picture}.

\section{\uppercase{Related Works}}
Recently, learning-based approaches have shown promising outcomes for diverse tasks of minimum-time planning and control. 
One of the learning-based approaches~\cite{song_2021_racing}, designed for drone racing scenarios, leverages deep reinforcement learning framework with additional privileged information for high-speed racing drones reaching speeds of $\sim$17m/s to navigate through gates autonomously. 
Similarly, \cite{robert_2023_agile_flight_cluttered} additionally include perception awareness for navigation in cluttered environments to optimize for flight efficiency and visual awareness. Perception awareness has also shown remarkable progress in drone racing~\cite{kaufmann2023champion} to achieve performance at the level of the best human pilots. 
However, these works have been only designed for specific race tracks with minimum perturbations (as seen in \cite{song_2021_racing}) and do not generalize for state-to-state agile flight. 

The prior methods are limited to a single task of navigation through race tracks. 
To address this limitation, \cite{Scaramuzza_2025_MTRL} presented a multi-Task DRL framework with a shared physical dynamics across tasks through shared actor and separate critic for each task. 
The authors provided a single policy that is capable of performing multiple tasks: stabilizing, tracking a racing trajectory, and velocity tracking, without the need for separate policies. Yet, without an explicit transition mechanism between different flight modes (e.g., switching between tracking a racing trajectory and stabilizing task mid-flight).

Learning-based control methods have also been shown to adapt to different quadrotor dynamics without manual tuning. 
The work presented in \cite{antonio_2023_near_hover} introduces a DRL based near-hover position controller that adjusts control gains in real-time that showed a rapid adaptation to quadrotors with mass variations up to 4.5 times along with disturbances such as external force, payloads or disparities in drone parameters. The authors of \cite{zhang2024learningbasedquadcoptercontrollerextreme} combined imitation learning (for a warm start) and reinforcement learning to fine-tune control in response to changing dynamics.
This work demonstrated adaptation to disturbances arising from an off-center payload
and external wind for quadrotors with mass differences and various propeller constants.

Deep reinforcement learning has also shown immense promise for landing a quadrotor. 
The authors of \cite{2021_inclined_landing} illustrated landing of a quadrotor on an inclined surface by using PPO with sparse rewards. 
Prior to this, \cite{2018_landing} presented the use of DRL for landing on ground markers with low resolution images, while \cite{2019_Landing_moving_platform} showed the possibility of landing on a moving platform.

The work presented in \cite{Eschmann_2024_learning_in_seconds} elucidated the feasibility of real-time learning by training an off-policy asymmetric actor-critic RL framework within 18s on a consumer-grade laptop with demonstration of successful real-world flight with minimal tuning. Previously, \cite{Hwangbo_2017_RL} successfully showed waypoint following using an approach based on Deterministic Policy Optimization (DPG). However, the aforementioned methods do not consider the minimum-time objective. Our proposed algorithm tackles this problem by designing a learning-based planner-controller optimized for minimizing time, capable of generalizing across varying start and goal positions. Our policy is not restricted to a single task, capable of seamlessly transitioning between waypoint flight and stable hovering. 

\section{\uppercase{Problem Formulation and Methodology}}
In this section we present the quadrotor dynamics considered during the training and deployment of our proposed method, formulate the control problem and explain the choice of observation and action spaces together with design of applied rewards. 
\subsection{System Dynamics}
The quadrotor is modeled with state $\boldsymbol{x}=[\boldsymbol{p}, \boldsymbol{q}, \boldsymbol{v}, \boldsymbol{\omega}, \boldsymbol{\Omega}]^{T}$ which consists of position $p \in \mathbb{R}^{3}$, velocity $\boldsymbol{v} \in \mathbb{R}^{3}$, unit quaternion rotation $\boldsymbol{q} \in \mathbb{S O}(3)$, body rates $\boldsymbol{\omega} \in \mathbb{R}^{3}$, and the rotors' rotational speed $\Omega$. The dynamics equations are

\begin{subequations}
\begin{align}
&\dot{\boldsymbol{p}}=\boldsymbol{v}\text{ ,}  \\
&\dot{\boldsymbol{q}}=\frac{1}{2} 
\boldsymbol{q} \odot\left[\begin{array}{c} 
0 \\
\boldsymbol{\omega}
\end{array}\right] \text{ ,}\\
&\dot{\boldsymbol{v}}=\frac{R(\boldsymbol{q})\left(\boldsymbol{f}_{T}+\boldsymbol{f}_{D}\right)}{m}+\boldsymbol{g} \text{ ,}\\
&\dot{\boldsymbol{\omega}}=\boldsymbol{J}^{-1}(\boldsymbol{\tau}-\boldsymbol{\omega} \times \boldsymbol{J} \boldsymbol{\omega})\text{ ,}
\end{align}
\end{subequations}

where $\odot$ denotes the quaternion multiplication, $R(\boldsymbol{q})$ is the quaternion rotation, $\boldsymbol{f}_{D}$ is the drag force vector in the body frame, $m$ is the mass, $\boldsymbol{J}$ is diagonal inertia matrix, and $\boldsymbol{g}$ denotes Earth's gravity. The speeds of the propellers $\boldsymbol{\Omega}$ are modelled as a first-order system, $\dot{\boldsymbol{\Omega}}=\frac{1}{k_{m o t}}\left(\boldsymbol{\Omega}_{c}-\boldsymbol{\Omega}\right)$ with $\boldsymbol{\Omega}_{c}$ being the commanded speed and $k_{m o t}$ the time constant.

The collective thrust $\boldsymbol{f}_{T}$ and torque $\boldsymbol{\tau}_{b}$ are calculated as:

\begin{align}
&\boldsymbol{f}_{T}=\left[\begin{array}{c}
0 \\
0 \\
\sum f_{i}
\end{array}\right]\text{,} \\
&\boldsymbol{\tau}=\left[\begin{array}{c}
l / \sqrt{2}\left(f_{1}-f_{2}-f_{3}+f_{4}\right) \\
l / \sqrt{2}\left(-f_{1}-f_{2}+f_{3}+f_{4}\right) \\
\kappa\left(f_{1}-f_{2}+f_{3}-f_{4}\right)
\end{array}\right] \text{,}
\end{align}

where $\kappa$ is the torque constant and $l$ is the arm length. Here, individual motor thrusts $f_{i}$ are functions of the motor speeds using the thrust coefficient $c_{f}$,

\begin{align}
f_{i}(\Omega)=\left[c_{f} \cdot \Omega^{2}\right]\text{.}
\end{align}

\subsection{Task Formulation}
The minimum-time planning and control problem is formulated in the reinforcement learning (RL) framework using a Markov Decision Process (MDP) represented as a tuple ($\mathcal{S}, \mathcal{A}, \mathcal{P}, \mathcal{R}, \gamma$). The action $a_t \in \mathcal{A}$ is executed at every time step $t$ to transition the agent from the current state $s_t \in \mathcal{S}$ to next state $s_{t+1}$ with the state transition probability $\mathcal{P}(s_{t+1}|s_t,a_t)$ receiving an immediate reward $r_t \in \mathcal{R}.$

The objective of deep reinforcement learning is to optimize the parameters $\theta$ of a neural network policy $\pi_\theta$ so that the learned policy maximizes the expected discounted reward. The objective function is formulated as 

\begin{equation}
\pi^*_\theta = \arg\max_{\pi} \mathbb{E} \left[ \sum_{t=0}^{\infty} \gamma^t r_t \right] \text{.}
\end{equation}

\subsection{Observation and Action Spaces}The observation space is defined as the vector $o_t = [v_t,\mathbf{R}_t,p_t,\omega_t,\phi_t, a_{t-1}] \in \mathbb{R}^{23}$ which corresponds to linear velocity of the quadrotor, rotation matrix,  relative position of the quadrotor to that of the goal position, input body rates around each axis of the quadrotor, relative heading of the quadrotor to that of the goal and the action executed at the previous time step.

The agent is such that the observation is mapped to the action consisting of collective thrust and body rates around each of the axes. This defines the action space to be $a_t = [\tau_t, \omega_{xt}^c,\omega_{yt}^c, \omega_{zt}^c]$.

\subsection{Reward Shaping} \label{sec:reward_shaping}
The objective of deep reinforcement learning is to maximize the cumulative discounted reward. To this end, our rewards consist of two main components, one direct, simple function-based rewards for position and velocity, and one capturing proxy reward via the usage of point-mass model (PMM)\cite{matej_2024_pmm} trajectories to consider the minimum-time state-to-state objective. 

\subsubsection{Direct Rewards} These rewards are designed to consider various aspects that are crucial for both hovering and the minimum-time state-to-state objective. In this paper, we consider total direct reward ($r_D$) to be the sum of the following components. 
\begin{itemize}
    \item \textbf{Distance to Goal}: $ r_g = K_g \left( 1 - \frac{\lVert p_t - p^g \rVert}{G} \right)$
    \item \textbf{Heading}: $ r_\phi = K_\phi| \phi_t - \phi_g |$
    \item \textbf{Stay at Goal}: $r_s = K_s(s_p \cdot s_g)$, where $s_p = \lVert p_t - p_{t-1} \rVert$, $s_g = \lVert p_t - p^g \rVert$ 
    \item \textbf{Acceleration}: $r_a = K_a \lVert a_t \rVert $ 
    \item \textbf{Angular Velocity}: $r_\omega = K_\omega \omega_t$
    \item \textbf{Thrust Smoothing}: $r_\tau = K_\tau| \tau_t - \tau_{t-1}|$
    \item \textbf{Commanded Angular Velocity}: $r_c = K_c|\omega^c_t - \omega^c_{t-1}|$,
\end{itemize}
where $ p \in \mathbb{R}^3, a_t \in \mathbb{R^3}, \phi_t \in [-\pi, \pi], \omega_t \in \mathbb{R^3}$ are the current position, acceleration, heading and angular acceleration of the UAV at time $ t $, $ p^g \in \mathbb{R}^3,\phi_g \in [-\pi, \pi] $ are the position and heading of the desired goal. $G$ is the acceptable convergence radius from the goal point. $\omega^c_t \in \mathbb{R}^3, \omega^c_{t-1} \in \mathbb{R}^3$ are the commanded velocities and $\tau_t, \tau_{t-1}$ are the commanded thrusts at times $t, t-1$. 
$K_g, K_\phi, K_s, K_a, K_\omega, K_\tau, K_c$ are the scaling factors.

The \emph{Distance to Goal} reward acts as an indicator for the quadrotor to move towards the goal, which is especially valuable during the exploration phase. 
\emph{Heading} reward awards if the quadrotor moves towards the heading goal and penalises it if not by a factor of the absolute difference between the current heading and the goal heading.
The \emph{Stay at Goal} reward keeps the quadrotor at the goal by rewarding if the movement direction vector of the quadrotor is aligned with the vector towards the goal. 
The \emph{Acceleration} and \emph{Angular Velocity} rewards penalize if either of these quantities are too high. 
These rewards are essential in finding the balance between stable hovering and agile start-to-goal flight. \change{We observed that, in the absence of these penalties, the quadrotor demonstrated effective agile start-to-goal flight but failed to maintain stable hovering. These reward components are crucial for enabling consistent multi-goal behavior.} 
The \emph{Thrust Smoothing} and \emph{Commanded Angular Velocity} rewards are temporal smoothing rewards. 
They penalise the quadrotor if the change is too high within immediate time steps of particular commanded values. 
This ensures the outputs of the policy do not contain any sharp perturbation which could cause instability during flight.

\begin{figure*}[h!]
	\centering
	\includegraphics[width=2.0\columnwidth]{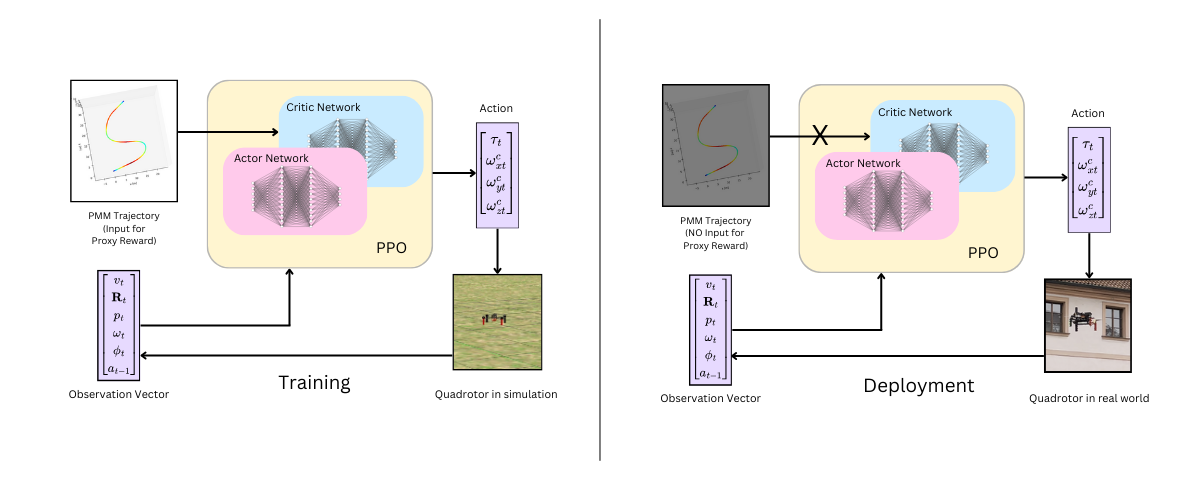}
	\caption{\textbf{System Overview}: We train the network with the trajectories generated through PMM planner as a proxy reward for minimum-time state-to-state objective. The network later does not have access to these trajectories during deployment. 
	}
    \label{fig:flowchart}
\end{figure*}


\subsubsection{Proxy Reward with PMM} In order to drive the minimum state-to-state objective, we use trajectories generated by the PMM planner~\cite{matej_2024_pmm} to compute proxy reward to closely approximate the true performance objective while providing feedback at every time step. The proxy reward is also instrumental in combining the planning and control objective for the learning process. The PMM planner planner generates the minimum-time trajectories over multiple waypoints within milliseconds by utilizing a point-mass model of the UAV combined with thrust decomposition algorithm that enables the UAV to utilize all of its collective  thrust.  

A time-optimal PMM trajectory segment can be described as:
\begin{equation}\label{eq:pmm_1seg}
	\begin{aligned}
		p_1 &= p_0 + v_0 t_1 + \dfrac{1}{2} a_1 t_1^2, \qquad
		v_1 = v_0 + a_1 t_1,\\
		p_2 &= p_1 + v_1 t_2 + \dfrac{1}{2} a_2 t_2^2, \qquad
		v_2 = v_1 + a_2 t_2,
	\end{aligned}
\end{equation}

where $ p_0, v_0 \in \mathbb{R} $ are the start position and velocity, $ p_2, v_2~\in~\mathbb{R} $ are the end position and velocity, $ t_1, t_2 \in \mathbb{R} $ are the time durations of the corresponding sub-segments of the trajectory and
\begin{equation}\label{eq:pmm_acc_limits}
	a_i \in \{a_{\text{min}}, a_{\text{max}}\}, \; i = 1,2, \; a_i \in \mathbb{R},
\end{equation}
are the optimal control inputs.
\mbox{$ a_1=a_2 $} is omitted as it is equivalent to $ t_1 = 0 $ or $ t_2 = 0 $. This omission leaves two two combinations of possible values for $ a_1 $ and $ a_2 $.

Analytical solutions to the equations \eqref{eq:pmm_1seg} can be derived, revealing a total of four possible solutions. For each combination of  $ a_1 $ and $ a_2 $ values, two solutions exist. The final solution is selected based on the criteria that all unknowns must be real numbers and that the real-world constraint of non-negative times,  $ t_1 \geq 0 $ and $ t_2 \geq 0 $, is met.

We sample a PMM trajectory generated between the initial and goal position. The proxy reward $r_{PMM}$ is defined as
\begin{equation}
    r_{PMM} = k_p^{PMM}\lVert p_t - p_t^{PMM} \rVert + k_v^{PMM}\lVert v_t - v_t^{PMM} \rVert
\end{equation}
where $ p_t, v_t $ describe the state of UAV at time $ t $ and $ p_t^{PMM}, v_t^{PMM} $ are position and velocity values sampled from PMM trajectory generated between the initial and goal position. $k_p^{PMM}, k_v^{PMM}$ are scaling factors. 

This reward incentivises the quadrotor to track the PMM trajectory akin to a linear MPC tracking the trajectory. 

\begin{table}[h]
    \centering
    \caption{Values of the rewards scaling parameters}
    \begin{tabular}{l c l c l c}
        \toprule
        $K_g$ & 0.2 & $K_\phi$ & -1.0 & $K_s$ & 0.2  \\
         $K_a$ & -0.15 &  $K_\omega$ & 0.25 &  $K_\tau$ & 0.4\\
          $K_c$ & 0.35 &  $k_p^{PMM}$ & -3.0 & $k_v^{PMM}$ & -0.3 \\
        \bottomrule
    \end{tabular}
    \label{table:scaling_params}
\end{table}
\vspace{-1em}

\section{\uppercase{Implementation Details}}
This section details the specifications of the simulation environment we used for training and evaluation, the details of the hardware equipment used for the real-world experiments along with the network architecture and implementation details of the proposed algorithm.

\subsection{Simulation Environment}
We use an in-house developed simulator for training and testing of our control policies. 
The inputs to our policy are linear velocity of the quadrotor, rotation matrix,  relative position of the quadrotor to that of the goal position, input body rates around each axis of the quadrotor, relative heading of the quadrotor to that of the goal, and the action executed at the previous time step. 
The state of the quadrotor (position, velocity, rotation matrix, linear and angular velocities) comes from the simulation platform and the reference PMM trajectories are generated at the simulated time point. 
The policy outputs mass-normalized collective thrust and the body rates. 
The maximum duration of each of the simulated environments is 5s with early termination if the quadrotor goes out of the bounds of the simulated environment. 
The simulation step size is 1ms and the control frequency is set to 100Hz.

We integrate the trained RL algorithm into the multirotor Unmanned Aerial Vehicle control and estimation system developed by the Multi-robot Systems Group (MRS) at the Czech Technical University (CTU) \cite{MRSBaca2021}, \cite{MRSHert2023}. 
For the simulated experiments, we use the existing multirotor dynamics simulation tools available in the system \cite{MRSHert2023}. 
\subsection{Hardware Specifications}
We utilize a custom-built agile quadrotor with a diagonal motor-to-motor span of 300 mm and a total weight of 1.2 kg. 
It is controlled by our open-source MRS system architecture \cite{MRSHert2023} operating on a Khadas Vim3 Pro single-board computer.
The quadrotor has a CubePilot Cube Orange+ flight controller with PX4 firmware acting as the low-level controller. 
It is powered by a 4S Lithium Polymer battery. 
We use Holybro F9P RTK GNSS module on the quadrotor for the state estimation. 
It receives real-time corrections from a base station, which are combined with the flight controller's IMU to generate odometry for the system. 

\subsection{Policy Training}
We use Proximal Policy Optimization (PPO) algorithm \cite{PPO}, a first order policy gradient method to train our agent. 
Although PPO is popular for its simplicity in implementation, our task is challenging due to the large dimensionality of the search space and the complexity of learning two distinctly different behaviours. Therefore, we emphasize important details that allowed us to achieve a policy capable of planning and control simultaneously.
\subsubsection{Network Architecture}
We represent the control policy with a deep neural network. We use PPO 
\cite{PPO} network architecture with 2 hidden layers with 256 ReLU nodes. The learning rate is linearly decreased starting from $5e-4$ ending at $2e-5$. Discount factor $\gamma$ is set to 0.99. The algorithm is set to collect 2000 experiences from each environment into the batch. Since parallelise the training with 500 environments the total batch size is 1 million.
Such a large batch size is needed to encapsulate the different behaviour in a large observation space.

\subsubsection{Parallelization} We trained our agent in simulation with 500 environments in parallel to increase the speed of the data collection process. Additionally, we were able to leverage this parallelization process to diversify the data collected by interacting with the environment. Since training requires significantly different environment interactions for flying in minimum-time between start and goal and hover at a setpoint, rollouts could be performed with multiple, diverse environments to learn a generalizing policy.

\subsubsection{Randomized Initialization and Dynamics} The training aims to learn simultaneously minimum-time planning and control. 
To this end, we randomize initial setpoint in position and heading to ensure exploration of necessary areas of state space. 
The position is randomly sampled within the interval \( [-20,20]m\) and the heading is randomly sampled in the interval \( [-\pi,\pi]\) radians. 

Furthermore, to ensure robustness, for each episode, the quadrotor's mass and inertia are randomized. 
These are used to imitate variations caused in the quadrotor parameters due to disturbances caused by environmental factors such as wind. 
The mass and inertia are both randomized by 30\%. 

\subsubsection{Curriculum Strategy} Having a large(40m) sampling range, that is essential for learning state-to-state generalizing behaviour, makes the problem incredibly challenging to learn. 
Therefore, we employ a curriculum approach to achieve the successful learning of desired objectives. 
The training is divided into 4 sub-tasks. 
The task difficulty increases with every stage in the curriculum. 
In the first stage, the start position (in the beginning of each episode) is sampled from the interval of \( [-1,1]m\) for each of $X, Y, Z$ axes. 
This ensures that the state space is small enough for successful exploration and helps the quadrotor learn the behaviour of moving towards the goal considering PMM trajectory as reference, which is given as proxy reward. 
The stage is stepped up once the RSME error calculated between the reached position after the rollout and the goal  
gets below 2 for 100 sampled trajectories. 
The position is sampled in the intervals of \( [-5,5]m\) ,\( [-10,10]m \) ,\( [-20,20]m\) for all the 3 axes in the second, third, and the fourth sub-tasks respectively. 
The heading is sampled from the interval \( [-\pi,\pi] \) radians for all the stages of the curriculum training.

\begin{figure}[htb!]
	\centering
	\includegraphics[width=0.5\textwidth]{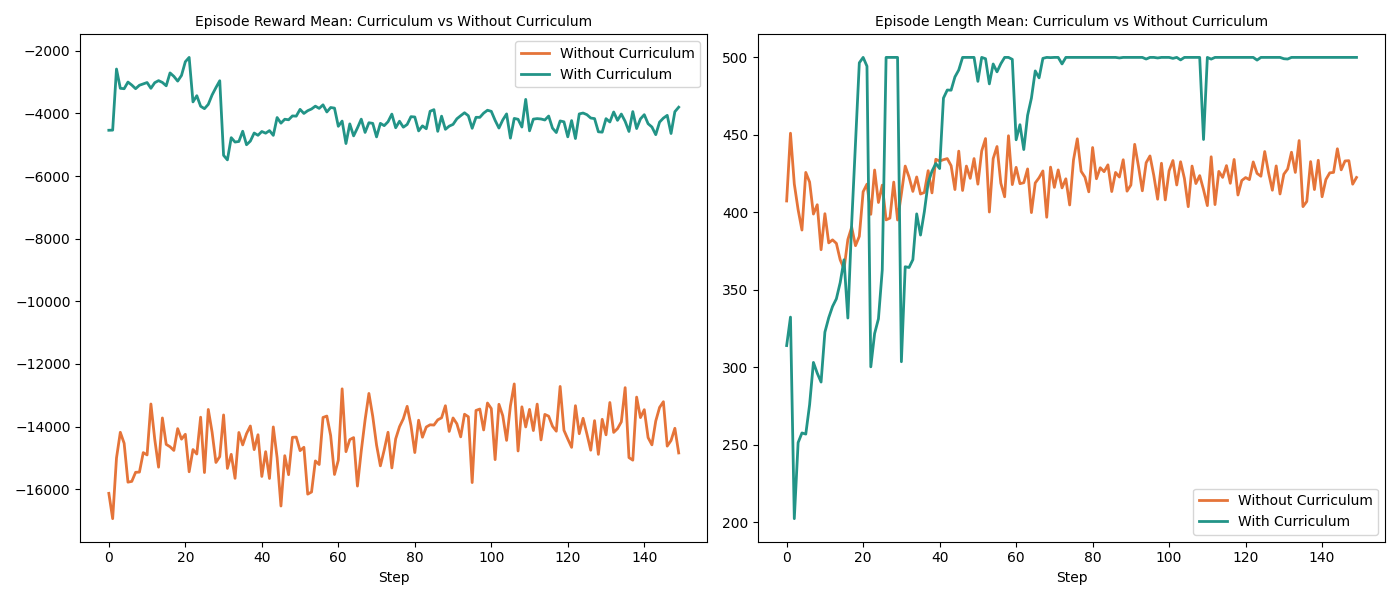}
	\caption{Comparison of mean rewards (left) and episode lengths (right) for policies trained with curriculum (green) and without curriculum (orange).}
	\label{fig:episode_length}
\end{figure}

\begin{figure}[ht!]
  \centering
  \subfloat [\textit{With curriculum}]{
    \includegraphics[width=0.46\linewidth]{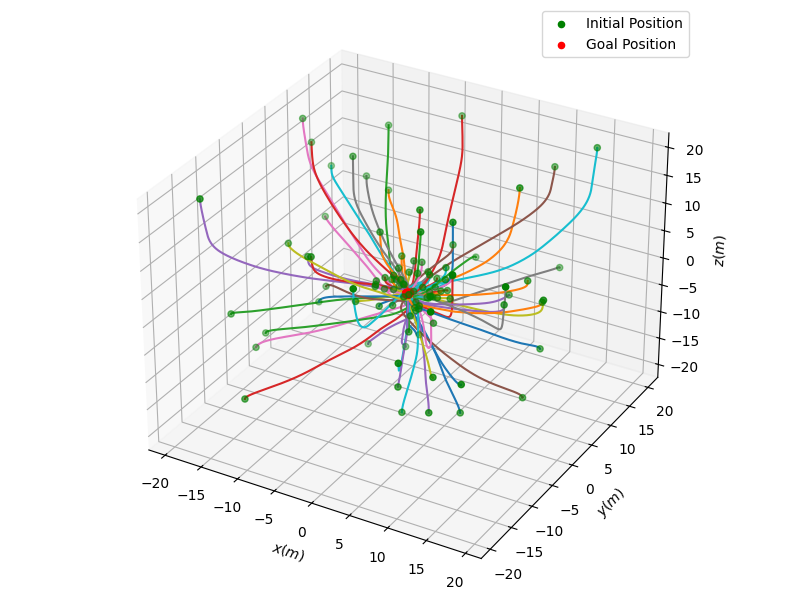}
    \label{fig:with_curriculum}
  }
  \subfloat[\textit{Without curriculum}]{
    \includegraphics[width=0.46\linewidth]{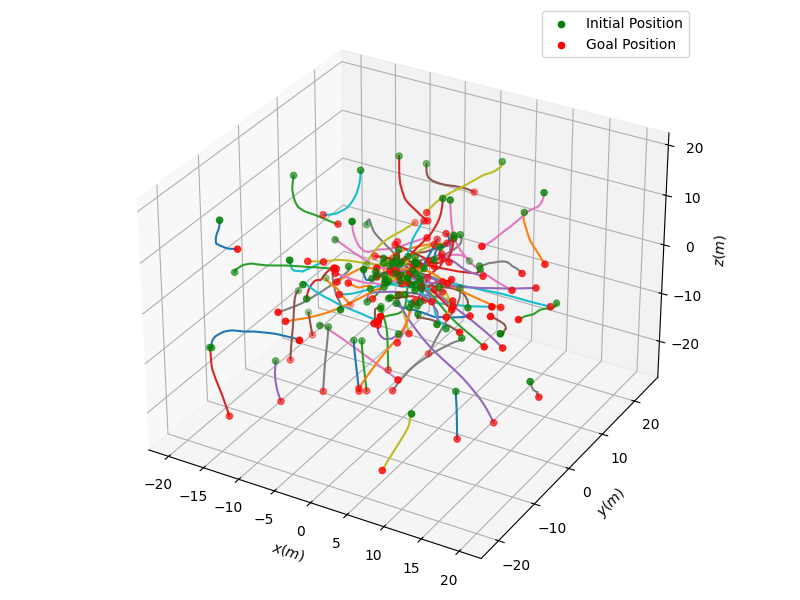}
    \label{fig:without_curriculum}
  }
  \caption{Visualization of typical desired quadcopter trajectories in 3D space during simulated tests with curriculum and without curriculum. 
  We sampled 100 initial positions. 
  The goal conditions of all trajectories are hovering at the origin.}
  \label{fig:curriculum_comaprision}
\end{figure}

\begin{table*}[ht!]
  \vspace{0.5em}
  \caption{Comparision of Flight Times and Maximum Velocity wrt baseline}
  \vspace{-0.3em}
  \centering
  \small
  \begin{adjustbox}{width=2.0\columnwidth,center}{
  \begin{tabular}{l| c c c | c c c}
    \toprule
    \multirow{2}{4em}{Trajectory}  & \multicolumn{3}{c|}{Flight time in seconds} &  \multicolumn{3}{c}{Maximum Velocity in m/s} \\   
    & Proposed RL & PMM + NMPC \cite{timeOptimal_NMPC} & PMM + NMPC w/ reduced velocity & Proposed RL & PMM + NMPC \cite{timeOptimal_NMPC}& PMM + NMPC w/ reduced velocity\\
    \midrule
    \textit{line} & 4.239 & 3.119 & 4.127 & 17.20 & 28.8 & 20.1 \\
    \textit{zigzag} & 7.924 & 5.092 & 7.503 & 14.5 & 22.6 & 15.9 \\
    \textit{3D semi-circle} & 8.169 & 5.614 & 8.202 & 12.8 & 19.0 & 12.5 \\
    \bottomrule
  \end{tabular}
  \label{tab:flight_time_comparision}
  \vspace*{-1.5em}}
  \end{adjustbox}
\end{table*}

\begin{figure*}[ht!]
	\centering
	\includegraphics[width=1.99\columnwidth]{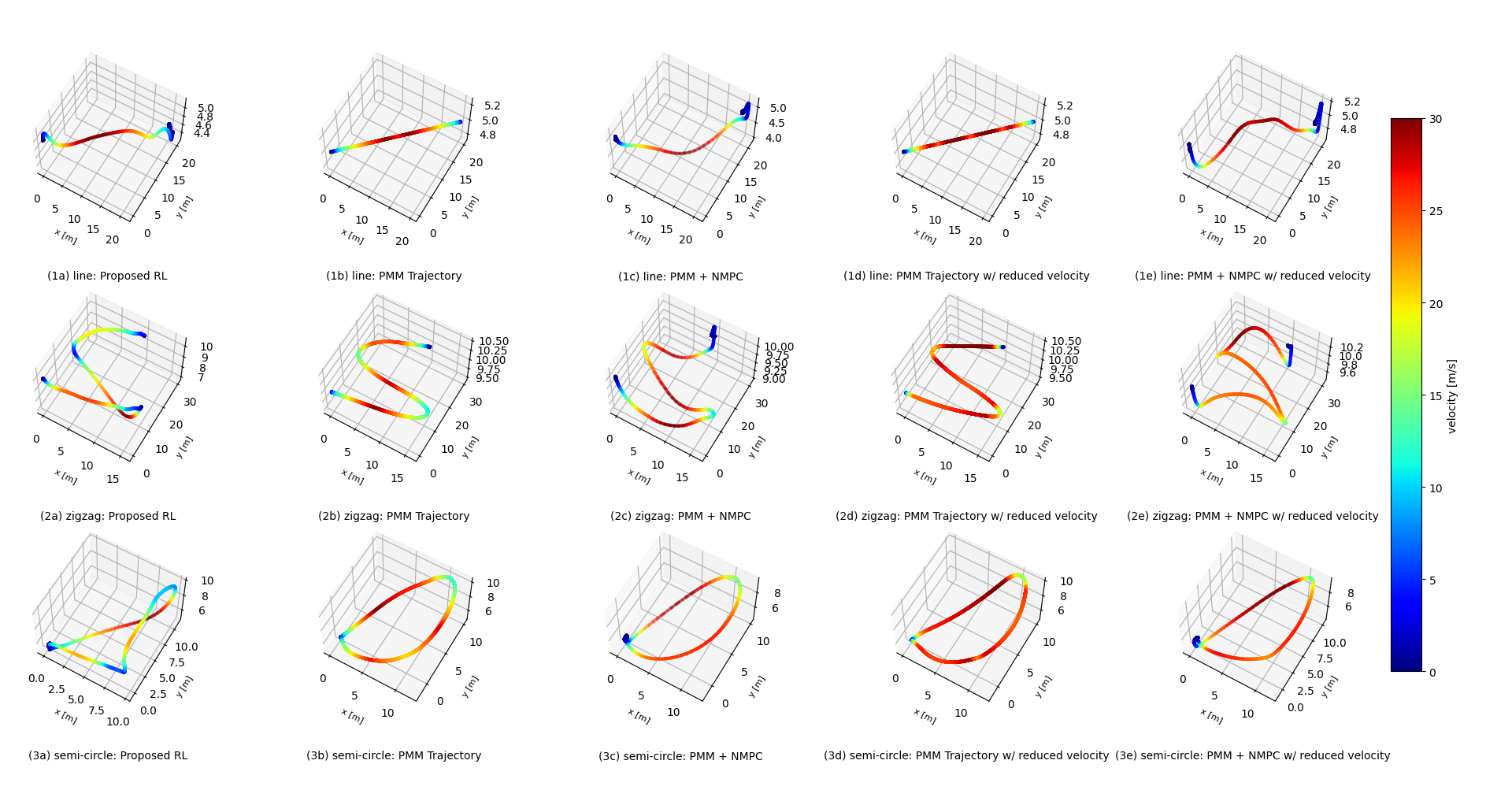}
    \vspace{-1em}
	\caption{\change{Velocity Profiles of proposed RL controller along with the generated PMM trajectories and NMPC tracking performance. The PMM trajectories with full and reduced velocities along with NMPC performance are shown.}}
	\label{fig:velocity_profiles}
    \vspace{-1em}
\end{figure*}

\section{\uppercase{Experimental Results}}
In this section, we outline the details of the experiments we conducted to validate our experiments. 
We begin by establishing the importance of the curriculum approach by evaluating the same objective learned with and without curriculum. 
Subsequently, we evaluate the performance of our method against a set of baseline methods in a controlled simulation environment. 
Eventually, we show the deployment of our controller on a real quadrotor (see Fig.~\ref{fig:intro_picture}) to validate the robustness of our approach.

\subsection{Simulation Experiments}
\subsubsection{Ablation Studies}: 
We compare the performance of the RL policy obtained when the training scheme contains curriculum and also when it is absent. To maintain consistency, we trained both of these polices with an objective to reach the same goal position when starting at a random initial position obtained from sampling a space spanning 40 x 40 x 40 m. 
Both the policies have access to proxy reward in the form of PMM trajectories during training. They are trained for 3000 million time steps with the hyperparameters detailed in section \ref{sec:reward_shaping} and scaling factors for rewards specified in Table \ref{table:scaling_params}. The training scheme used is the only sole distinction between the polices.

The results of the ablation studies are visualized through Figures \ref{fig:episode_length} and \ref{fig:curriculum_comaprision}. The learning approach with curriculum converges compared to training solely with the conventional approach. It could be observed from Fig.~\ref{fig:episode_length} that both the mean cumulative discounted reward as well as the mean episode length are higher with the use of the curriculum approach, substantiating the convergence of the proposed approach. 
This is more evident from the visualizations of 100 rollouts with each of the approaches shown in Fig.~\ref{fig:curriculum_comaprision}. 
The policy with the curriculum training paradigm successfully learns the objective while conventional RL fails to do so. 

\subsubsection{Comparison with Baseline}
We choose 3 trajectories: line, zigzag, and slanted semicircle as shown in Fig. \ref{fig:velocity_profiles} to evaluate the performance against the Nonlinear Model Predictive Controller (NMPC) \cite{timeOptimal_NMPC}  tracking the trajectory generated by the PMM planner. \change{Identical waypoints were provided to both the PMM planner and the proposed RL-based planner-controller. Subsequently, the trajectory generated by the PMM planner—tracked by an NMPC controller—and that produced directly by the proposed RL planner-controller are evaluated.} 
These curves together encapsulate the common manoeuvrers a UAV is expected to execute in the majority of applications.

\begin{figure}[ht!]
	\centering
	\includegraphics[width=0.5\textwidth]{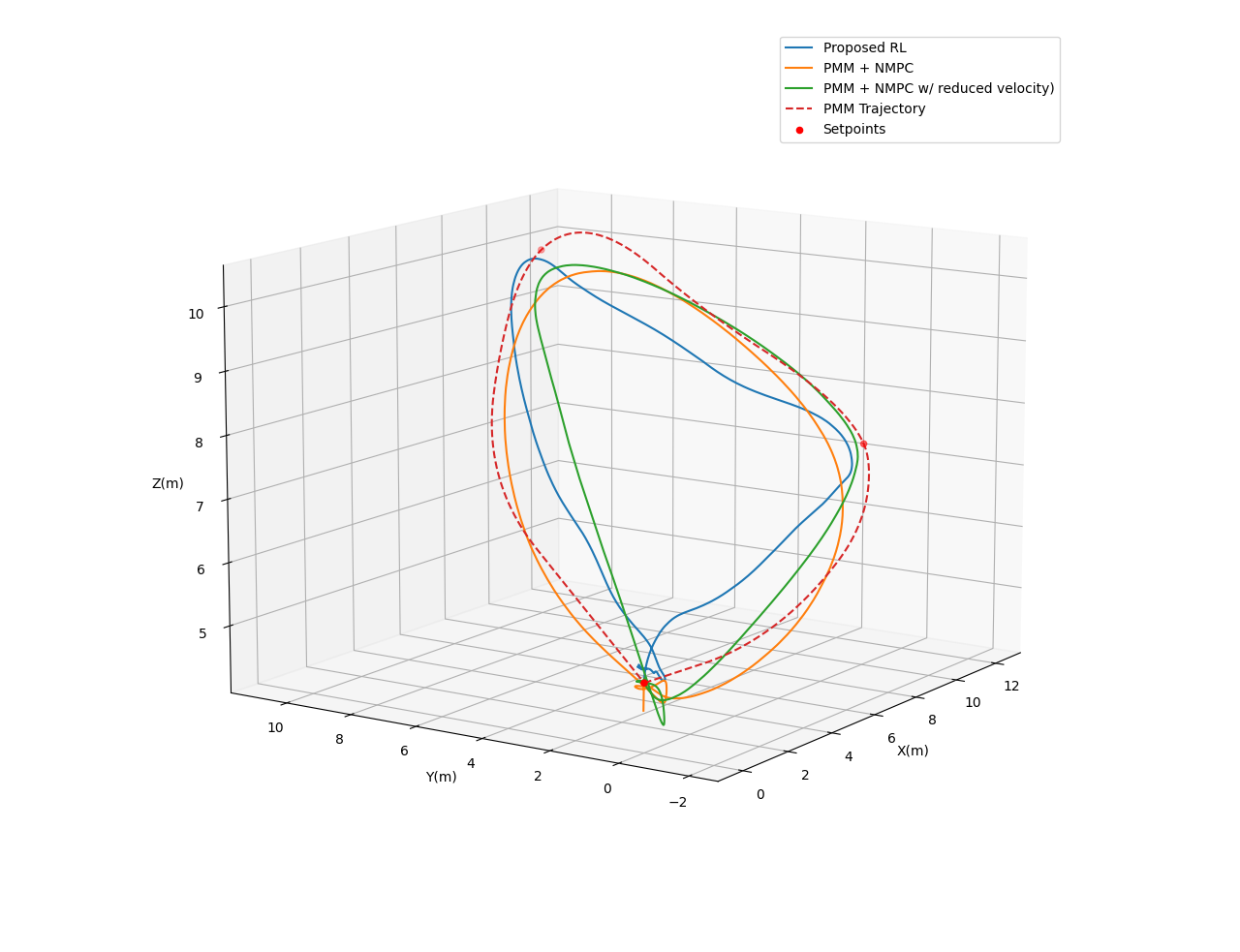}
        \vspace{-3.5em}
	\caption{\change{Comparison of path of the proposed RL algorithm with the baseline for the waypoints of a semi-circle.}}
	\label{fig:semi-circle}
    \vspace{-1em}
\end{figure}

\begin{figure}[ht!]
  \centering
  \subfloat [\textit{line}]{
    \includegraphics[width=0.46\linewidth]{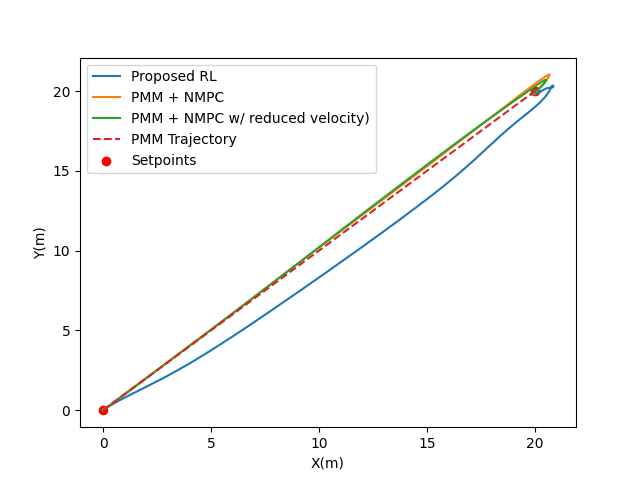}
    \label{fig:line}
  }
  \subfloat[\textit{zigzag}]{
    \includegraphics[width=0.46\linewidth]{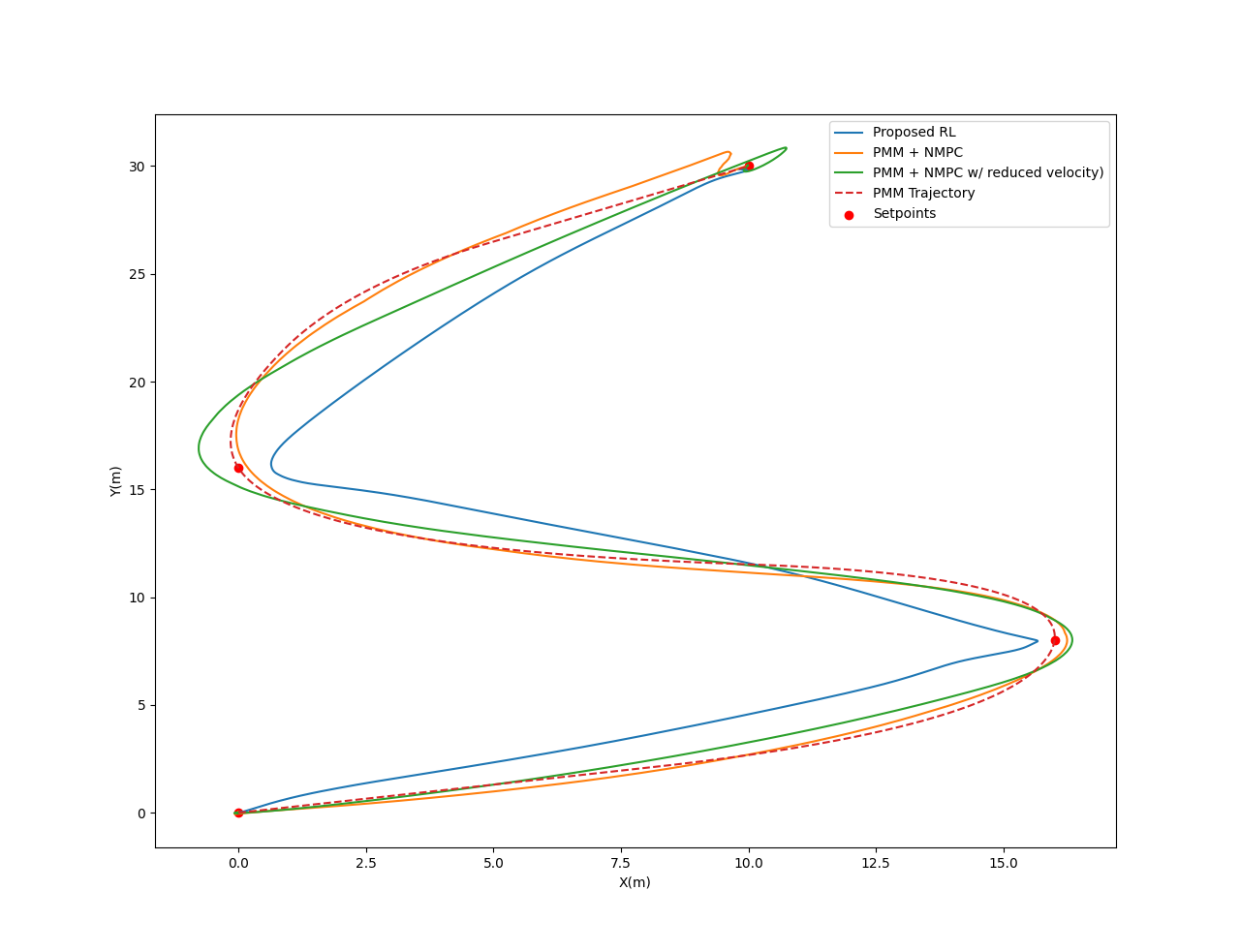}
    \label{fig:zigzag}}
  \caption{\change{Comparison of path of the proposed RL algorithm with the baseline for the waypoints of line and zigzag curve.}}
  \label{fig:sim_line_zz}
\end{figure}

Despite providing the same velocity and acceleration constraints to the PMM during training of the proposed RL algorithm and for the NMPC controller, we observe a significant loss in performance using the proposed algorithm. 
The proposed RL algorithm has to learn diametrically opposite behaviours of hovering and agile flight over a large distance. 
Therefore, the proposed RL algorithm learns to accommodate the opposing behaviours and realize a stable behaviour with smaller speed. 
To further substantiate this, we compared the performance of the proposed RL controller to NMPC tracking the PMM trajectories with reduced velocity. The velocity profiles for the trajectories flown are shown in Fig. \ref{fig:velocity_profiles}. 
When we matched the maximum velocity of the PMM trajectory to that of the proposed RL controller, the flight times are comparable as seen in Table \ref{tab:flight_time_comparision}. 
It is also interesting to note that since the proposed RL controller was trained on state-to-state trajectories, the path it takes between two waypoints is straighter when compared to the trajectory generated by the PMM planner, as shown in the Figures~\ref{fig:semi-circle} and \ref{fig:sim_line_zz}.

\subsection{Real World Experiments}
\begin{figure}[ht!]
  \centering
  \subfloat [\textit{Run 1}]{
    \includegraphics[width=0.46\linewidth]{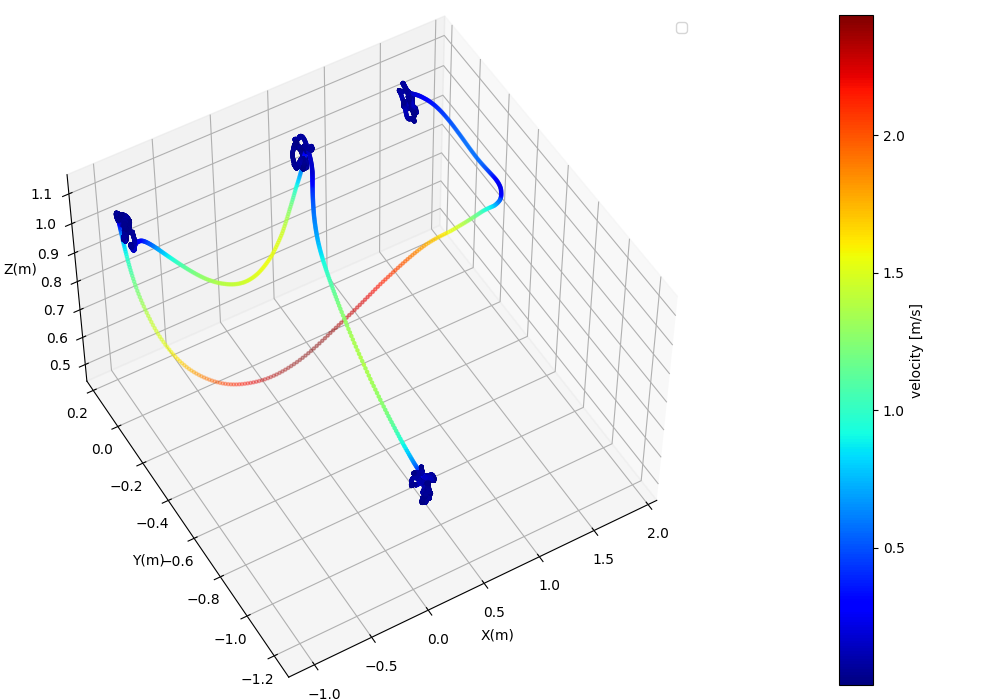}
    \label{fig:traj1}
  }
  \subfloat[\textit{Run 2}]{
    \includegraphics[width=0.46\linewidth]{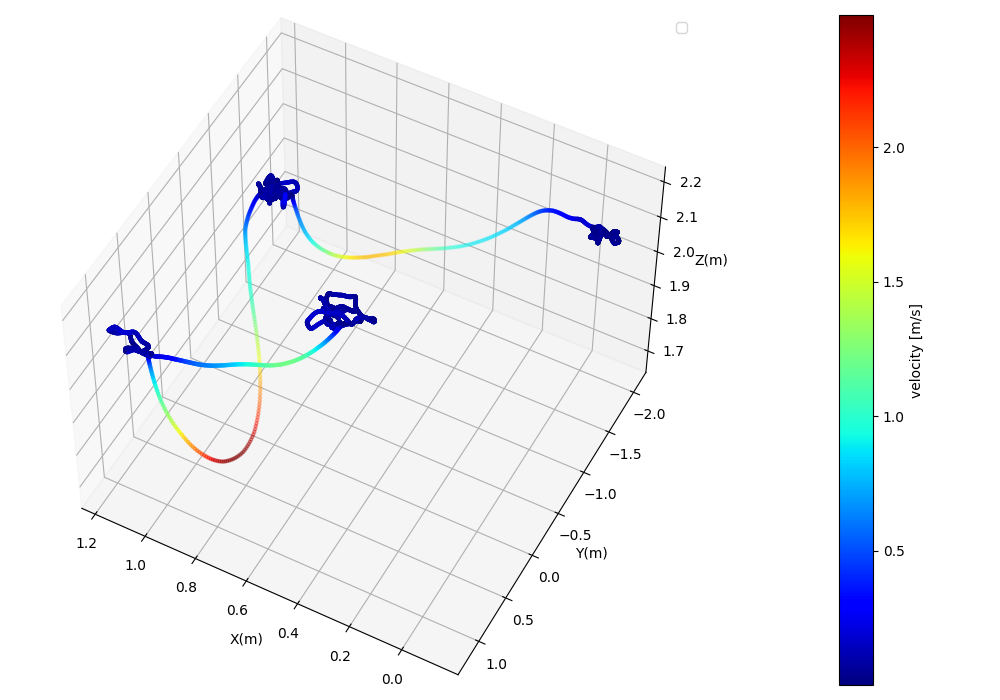}
    \label{fig:traj2}}
  \caption{\change{Performance of the proposed RL planner-controller between arbitrary way points in the real world.}}
  \label{fig:real_world}
\end{figure}

After verifying the proposed solution in simulation, we conducted experiments in the real world. 
We gave random goal waypoints to our controller to test the robustness of the generalization of the proposed controller. 
Due to constraints in space with the outdoor testing environment, we were only able to input goal points maximally 2m far. 
Therefore, we were only able to reach a maximum speed of 2.5 m/s. 

Our proposed solution was able to hover stably and reach the given way points successfully in two different runs as shown in Fig. \ref{fig:real_world} which showcases an ability of sim-to-real transfer. On the other hand, we observe loss of altitude initially. We believe this is due to the aggressive nature of the policy and mismatch in the PID control coefficients in the lower-level control and communication delays between the simulated environment and the real-world system. This shows the ability of our controller to be robust against the mismatches, which is a consequence of artificially introducing system mismatches by randomizing the mass, inertia and gravity coefficient during training.

\section{\uppercase{Conclusion}}
In this work we introduced a reinforcement learning framework that enables quadrotors to perform generalizable, minimum-time flights between any given start and goal states. Unlike traditional methods in autonomous drone racing that rely on predefined track layouts, this approach focuses on balancing agile flight with stable hovering. ff
The key innovation lies in using Point Mass Model (PMM) trajectories as proxy rewards to approximate the optimal flight objective. Additionally, curriculum learning is employed to enhance training efficiency and generalization. The proposed method is validated through simulations, where it is compared against Nonlinear Model Predictive Control (NMPC) tracking PMM-generated trajectories. Ablation studies further highlight the benefits of curriculum learning. Finally, real-world outdoor experiments demonstrate the robustness and adaptability of the learned policy, confirming its effectiveness in various challenging scenarios. In the future, we would like to conduct experiments over longer distances to better understand the potential of the approach and its full transferability to the real world.

\section*{Acknowledgment}
The authors thank Ond\v{r}ej Proch\'{a}zka for helping with the real-world experiments. This work  has been supported by the Czech Science Foundation (GAČR) under research project No. 23-06162M, by the European Union under the project Robotics and Advanced Industrial Production (reg. no. CZ.02.01.01/00/22\_008/0004590) and by CTU grant no SGS23/177/OHK3/3T/13.

\bibliography{ref.bib}{}
\bibliographystyle{IEEEtran}

\end{document}